\documentclass[11pt]{article}
\usepackage{konvens2019}
\usepackage{mathptmx}
\usepackage[scaled=.90]{helvet}
\usepackage{courier}
\usepackage{svg}
\usepackage{url}
\usepackage{latexsym}

\usepackage{algpseudocode}
\usepackage{algorithm}
\newcommand\CONDITION[2]%
  {\begin{tabular}[t]{@{}l@{}l@{}}
     #1&#2
   \end{tabular}%
  }
\algdef{SE}[WHILE]{While}{EndWhile}[1]%
  {\algorithmicwhile\ \CONDITION{#1}{\ \algorithmicdo}}%
  {\algorithmicend\ \algorithmicwhile}
\algdef{SE}[FOR]{For}{EndFor}[1]%
  {\algorithmicfor\ \CONDITION{#1}{\ \algorithmicdo}}%
  {\algorithmicend\ \algorithmicfor}
\algdef{S}[FOR]{ForAll}[1]%
  {\algorithmicforall\ \CONDITION{#1}{\ \algorithmicdo}}
\algdef{SE}[REPEAT]{Repeat}{Until}{\algorithmicrepeat}[1]%
  {\algorithmicuntil\ \CONDITION{#1}{}}
\algdef{SE}[IF]{If}{EndIf}[1]%
  {\algorithmicif\ \CONDITION{#1}{\ \algorithmicthen}}%
  {\algorithmicend\ \algorithmicif}%
\algdef{C}[IF]{IF}{ElsIf}[1]%
  {\algorithmicelse\ \algorithmicif\ \CONDITION{#1}{\ \algorithmicthen}}
  
\usepackage{enumitem}
\usepackage{booktabs}
\usepackage{graphicx}
\usepackage{balance}

\title{Teacher-Student Learning Paradigm for Tri-training: An Efficient Method for Unlabeled Data Exploitation}

\author{Yash Bhalgat\\
  Qualcomm AI Research \\
  {\tt  yashbhalgat95@gmail.com } \\
  \\
 \textbf{Zhe Liu, Pritam Gundecha, Jalal Mahmud, Amita Misra}  \\
  IBM-Research, Almaden \\
  {\{\tt liuzh, psgundec, jumahmud, amita.misra1\}@us.ibm.com} \\}
\date{}

\begin{document}
\maketitle
\begin{abstract}
Given that labeled data is expensive to obtain in real-world scenarios, many semi-supervised algorithms have explored the task of exploitation of unlabeled data. Traditional tri-training algorithm and tri-training with disagreement have shown promise in tasks where labeled data is limited. In this work, we introduce a new paradigm for tri-training, mimicking the real world teacher-student learning process. We show that the adaptive teacher-student thresholds used in the proposed method provide more control over the learning process with higher label quality. We perform evaluation on SemEval sentiment analysis task and provide comprehensive comparisons over experimental settings containing varied labeled versus unlabeled data rates. Experimental results show that our method outperforms other strong semi-supervised baselines, while requiring less number of labeled training samples.
\end{abstract}

\section{Introduction}

Machine learning algorithms often require large amount of labeled data for training. As collecting labeled examples can be expensive, semi-supervised learning has been proposed \cite{zhu2006semi}. Among the existing semi-supervised approaches, self-training \cite{triguero2015self}, co-training \cite{blum1998combining}, and tri-training \cite{zhou2005tri} are the most notable ones. However, they suffer from one major issue of the gradually increased level of noise during the iterative labeling process. This problem can be attributed to two factors: (1) static labeling threshold, and (2) inappropriate stopping criteria.  

Many self-labeled algorithms iteratively enlarge labeled training set with unlabeled instances whose prediction confidence is larger than a static labeling threshold. Static labeling threshold produces a good classification performance only when the proportion of correctly labeled instances remains above a constant level. However, given the continuously added noisy labels during the semi-supervised process \cite{triguero2015self}, it is unlikely that any fixed assignment of the threshold will produce optimal classifications.

Besides, deciding when to stop the iterative instance labeling process is also critical for the self-labeled techniques. Existing stopping criteria include: setting a threshold on the number of labels that the algorithm is willing to generate, or stopping the labeling process when little to no accuracy increase occurs in an iteration. Stopping criteria is still an open issue, as too conservative or too liberal stopping criteria may produce many mislabeled examples to the self-labeled process.

To solve the two challenges, we propose a new tri-training-based method, called tri-training with teacher-student paradigm. Specifically, in each iteration, a double-teacher-single-student teaching relation is established based on predefined teacher and student thresholds, where teachers teach the student with generated proxy labels on the unlabelled data. Along the teaching process, the teacher-student relationship is continuously adjusted with adaptive teacher and student thresholds. The teacher-student relationship terminates on either running out of teachable instances or when reaching a \textit{graduation point}, where the student threshold equals the teacher threshold. 

We evaluate the tri-training with teacher-student paradigm approach on the sentiment analysis task of SemEval-2016 over various labeled-unlabeled data ratios. The proposed method outperforms many strong baselines in terms of gaining better prediction performances while consuming less number of unlabeled examples. 

\section{Method}
Assume we are given a set of unlabeled samples $U$ as well as a set of labeled samples $L$, where $L$ $\ll$ $U$. The proposed method starts by training three independent base classifiers $m_i$, $m_j$, $m_k$ on bootstrapped sample subsets $S_i$, $S_j$, $S_k$ respectively taken from $L$. The aim of the bootstrap sampling is to increase the diversity of base classifiers trained through the labeled set. Next, for every sample $x$ in $U$, each of the trained models $m_i$, $m_j$, $m_k$ predicts a label $c_i$, $c_j$, $c_k$ with corresponding prediction probability $p_i(c_i|x)$, $p_j(c_j|x)$, $p_k(c_k|x)$. 

\subsection{Teacher-Student Assignment}
Instead of assigning $x$ a majority voted label, as implemented in the original tri-training \cite{zhou2005tri}, here we model the learning task from a teacher-student perspective. In each iteration of our proposed approach, two classifiers ($m_j$ and $m_k$) are ascertained to be teachers if their prediction probabilities $p_j(c_j|x)$ and $p_k(c_k|x)$ are both larger than the teacher threshold $\tau_t$. The other classifier $m_i$ is then treated as student if its prediction probability is less than the student threshold $\tau_s$. An unlabeled sample $x$ in $U$ will only be assigned a label after it is identified as \textit{teachable}. Teachable examples are defined according to the function \textit{SelectTeachableSamples}, as shown in Algorithm \ref{shortlisting}. The required criteria are as follows: Firstly, the predicted labels $c_j$ and $c_k$ from the two teachers $m_j$ and $m_k$ must agree with each other. Second, both teachers' prediction confidences $p_j$ and $p_k$ must exceed $\tau_t$ and at the same time, the student's confidence $p_i$ must be less than $\tau_s$. This setting of using two teachers ensures that bias in any of these models doesn't affect the quality of the information taught to the student. It's similar to the real-life teacher-student learning process, where only qualified teachers can teach students things that they are the most comfortable with. 
Here, it is important to note that the teacher-student roles are rotated in each iteration, $i\in\{1,2,3\}, (j,k\neq i)$, allowing each classifier to learn from the other classifiers' experiences, as $m_i$ is further trained with the original labeled set $L$ along with the identified teachable samples $L_i$. 



\subsection{Adaptive Thresholds} Another novel aspect that we adopt from real-world teaching scenarios to the proposed method is the continuously adjusted teacher-student relationship. To be more specific, as a student learns from the teachers, it would become more confident of its prior knowledge taught by the teachers. In that sense, the student threshold $\tau_s$ increases monotonically in every iteration. On the other hand, as student progresses through the learning process, the teachers are supposed to teach them more advanced cases, i.e. cases where the teachers are less confident about. This is captured in our approach by monotonically decreasing the teacher threshold $\tau_t$. For this work, we chose a linear adaptive rate for the adaptive process as shown in line 10 and 11 of Algorithm \ref{teacher-student}.
\begin{algorithm}[t]
\small
\caption{Teacher Student Tri-training}\label{teacher-student}
\begin{algorithmic}[1]
\Require $L$ - set of labeled samples, $U$ - set of unlabeled samples, $m_{i,j,k}$ - teacher-student models, $\tau_t$ - teacher threshold, $\tau_s$ - student threshold, $\lambda_t, \lambda_s$ - teacher-student adaptive rates
  \For {$i\in \{1..3\}$}
      \State $S_i\gets bootstrap\_sample(L)$
      \State $m_i\gets train\_model(S_i)$
  \EndFor

  \While{$\tau_s \leq \tau_t$}
      \For{$i\in \{1..3\}$}
          \State $L_i \gets SelectTeachableSamples(U, m_{i,j,k}, \tau_t, \tau_s)$
          \State $m_i\gets train\_model(L \cup L_i)$
      \EndFor
      \State $\tau_t \gets \tau_t - \lambda_{t}$
      \State $\tau_s \gets \tau_s + \lambda_{s}$
  \EndWhile
  \State apply majority vote over $m_{i,j,k}$
\end{algorithmic}
\end{algorithm}

\begin{algorithm}[t]
\small
\caption{Select Teachable Samples}\label{shortlisting}
\begin{algorithmic}[1]
\Require $U$ - set of unlabeled samples, $\tau_t$ - teacher threshold, $\tau_s$ - student threshold, $m_i$ - student model, $m_{j,k}$ - teacher models
	\State $\pi \gets \emptyset$
	\ForAll{$x \in U$}
    	\If{$c_j=c_k$}
        	\State $tcf = \min(p_j(c_j|x), \; p_k(c_k|x))$
            \State $scf = p_i(c_j|x)$
            \If{$\; tcf> \tau_t \;\; \& \;\; scf < \tau_s \;$}
            	\State $\pi \gets \pi \cup \{(x,c_j(x))\}$
            \EndIf
        \EndIf
    \EndFor
    \State \Return{$\pi$}
    
\end{algorithmic}
\end{algorithm}
\subsection{Stopping Criteria}
\label{sect:stopping}
Existing self-labeled techniques often stop when no sample can be labeled, or no performance improvement occurs in an iteration. The original tri-training paper introduces an error constraint that checks if a peak performance has been reached. However, the error measurement is conducted only on the labeled dataset, hence assuming that the labeled set distribution is representative of the unlabeled set distribution. Tri-training may also lead to a limited number of co-labeling examples for training and a premature termination while dealing with large datasets \cite{chou2016boosted}.

In this work, we present our stopping criterion by comparing the student's confidence threshold with the teacher's threshold during each training iteration. We assume that when a student reaches the same confidence level as the teachers in a particular iteration, then there is nothing to be learned for the students from the teachers. This happens in our algorithm \ref{shortlisting}, when $\tau_s$ $\geq$ $\tau_t$. At this point, adding newer samples to the training set of $m_i$ (the student) would not contribute to its learning anymore. In that sense, we called the point when $\tau_s$ $\geq$ $\tau_t$ as the \textit{graduation point}, so as to stop the tri-training process naturally when the constraint is reached.

\section{Evaluation}
\label{ssec:experiments}

\subsection{Experimental Settings}

\noindent{\bf Datasets.}  
We evaluate our model on the sentiment classification dataset of SemEval-2016 Task 4 Subtask A \cite{nakov2016semeval}. In total, there are 6000 training sentences, including 3094 positive, 863 neutral, and 2043 negative instances. We use 2000 sentences from the dev set for validation and we have 20632 for test. To test the model's generalizability, we subsequently examine it under different proportions of labeled data. We select 10\%, 20\%, 30\% and 40\% of the training set randomly as labeled samples $L$ and treat the rest as unlabeled $U$ by hiding their labels. Hidden labels are used later for quality check of the generated proxy labels.


\noindent{\bf Baselines.}
Since our method improves upon the foundations laid by the typical semi-supervised methods as mentioned in the related work section (e.g. tri-training and self-training), we compare with the following baselines:
\begin{enumerate}[leftmargin=0.9cm, itemsep=-0.3em]
\item \textit{NB STr} - Self-training with Naive Bayes as base classifier.
\item \textit{SVM STr} - Self-training with SVM as base classifier.
\item \textit{MLP STr} - Self-training with neural networks (multilayer perceptrons) as base classifier.
\item \textit{Tri} - Tri-training with SVM as base classifiers.
\item \textit{Tri-D} - Tri-training with disagreement with SVM as base classifiers \cite{sogaard2010simple}.
\end{enumerate}

Our proposed approach is tri-training with teacher-student paradigm (Tri-TS). We don't compare with co-training here because there are no clear independent views \cite{zhou2005tri} in the sentiment analysis task. We do not use any deep learning model as base learner in this study, as deep learning models may not perform well in the presence of limited labeled data.  We did try FastText \cite{joulin2017bag} as a proof case, but even under the $40\%$ label rate, its performance is unsatisfactory (an initial $F_1^{PN}$ of $0.346$ with an improvement of $+0.034$ using the proposed model). 

In all the baselines, we experiment with different base classifiers and their combinations, namely Naive Bayes, SVM and Neural Networks. We use a linear kernel (LinearSVC) for SVM. For the neural networks (MLP), we use 50 neurons in the hidden layer with a softmax output. We use Glove 300-dimensional word embeddings \citename{pennington2014glove}.\ After text-cleaning and tokenization, we average the word-embeddings for the tokens present in the sentence to get the feature vectors. For both the tri-training baselines, Tri and Tri-D, we obtain the best results with SVM as base classifiers. Hence, we report these for comparison with our approach.

Note that, as mentioned in Section \ref{sect:stopping}, for the baselines \textit{Tri} and \textit{Tri-D}, we use their own respective stopping criteria during evaluation, as a comparison to our newly proposed stopping criterion.






\noindent{\bf Parameter Tuning.}
All parameters required in both the proposed method and the baselines are fine-tuned using the validation set. A grid search is used to determine those parameter values that maximize each model's performance. For the proposed method $\tau_t$ is tuned $\in [0.7,1.0]$, $\tau_s$ $\in [0.6, 0.95]$. The best performed rates of $\lambda_t$ and $\lambda_s$ are found empirically as 0.001. For the tri-training baselines, we try to tune the error constraint as suggested in the original paper, but it generates only small number of proxy labels during the training process and terminates after very limited number of iterations. In that sense, we discard the error constraint and try the threshold based tri-training method as adopted in \cite{ruder2017knowledge} and \cite{sogaard2010simple}. Best performed parameters are obtained again via evaluations on validation set.
\subsection{Results}
\label{sect:pdf}
We evaluate our approach and the baselines from three different aspects: the overall model performance, the quality of generated proxy-labels, and the quantity of unlabeled data consumed. Model performances are reported using $F_1^{PN}$-score as adopted in the SemEval competition. 

\noindent{\bf Overall Performance.} 
The methods \textit{Tri} and \textit{Tri-D} both use majority voting to combine the three classifiers. For a fair comparison with these methods, after the training is completed, we perform majority voting on the test set to get the final predictions. In Table \ref{table:1}, we see that the proposed tri-training with teacher-student paradigm consistently outperforms the other baselines with higher prediction performance across different labeled versus unlabeled settings. The proposed method reaches a $F_1^{PN}$ of 0.523 using just 40\% of the labeled data, whereas the upper bound $F_1^{PN}$ is only 0.585, if the we train the base SVM classifier on the 100\% training dataset.


To better understand the effectiveness of the proposed teacher-student paradigm, we further look into the performance of each individual base classifier before the majority voting step,  We found that under the 10\% label rate, the maximum $F_1^{PN}$ achieved between the base classifiers and the final ensemble model was only 0.011, and such difference decreased to 0.005, when label rate increased to 40\%, which indicates indicates good quality of the base classifiers even without the ensemble step. In addition, same conclusion can also be inferred as the base classifiers in Tri-TS before ensemble performed better than the base classifiers in all the other baselines.

\begin{table}[t]
\begin{center}
\small
\begin{tabular}{ccccc}
\toprule


\textbf{} & \textbf{10\%} & \textbf{20\%} & \textbf{30\%} & \textbf{40\%}\\

\midrule

NB STr & 0.461 & 0.471 & 0.484 & 0.495\\
SVM STr & 0.465 & 0.469 & 0.478 & 0.489\\
MLP STr & 0.471 & 0.481 & 0.497 & 0.499 \\
Tri & 0.478 & 0.489 & 0.501 & 0.505\\
Tri-D & 0.485 & 0.499 & 0.507 & 0.511\\
\textbf{Tri-TS} & \textbf{0.498} & \textbf{0.507} & \textbf{0.519} & \textbf{0.523}\\

\bottomrule\\
\end{tabular}








\caption{$F_1^{PN}$ comparison averaged over 5 runs for different proportions of labeled data.
\label{table:1}}
\vspace{-5mm}
\end{center}\end{table}


\noindent{\bf Quality of Proxy-labels.} 
The quality of the assigned proxy-labels to the unlabelled data in each iteration determines how well the model learns. So, here, we evaluate the quality of all produced proxy-labels during the self-labeling process against the hidden ground truth to determine the effectiveness of the algorithms in terms of teaching the correct labels. Table \ref{table:2} shows that teacher models in our proposed method consistently produce high quality proxy-labels (88.93\% match with the hided ground truth labels) for the student model to learn. The other baselines tend to suffer from the problem of adding unreliable labels to the labeled dataset. We view this result as a confirmation of the usefulness of the adaptive threshold in terms of producing high quality proxy-labels on the unlabeled data.

\begin{table}[t]
\begin{center}
\small
\begin{tabular}{ccccc}
\toprule


\textbf{} & \textbf{10\%} & \textbf{20\%} & \textbf{30\%} & \textbf{40\%}\\

\midrule

NB STr & 65.81 & 67.14 & 63.36 & 70.15\\
SVM STr & 68.15 & 67.59 & 71.08 & 68.13\\
MLP STr & 76.81 & 77.71 & 79.07 & 78.29 \\
Tri & 71.78 & 76.49 & 75.71 & 73.35\\
Tri-D & 75.28 & 70.19 & 72.37 & 77.11\\
\textbf{Tri-TS} & \textbf{86.18} & \textbf{84.57} & \textbf{88.19} & \textbf{88.93}\\

\bottomrule\\
\end{tabular}

\caption{Percentage of matches between the produced proxy-labels and the ground truth averaged over 5 runs for different proportions of labeled data.
\label{table:2}}
\vspace{-5mm}
\end{center}\end{table}

\begin{figure}[!htbp]
    \centering
    \includegraphics[width=0.48\textwidth]{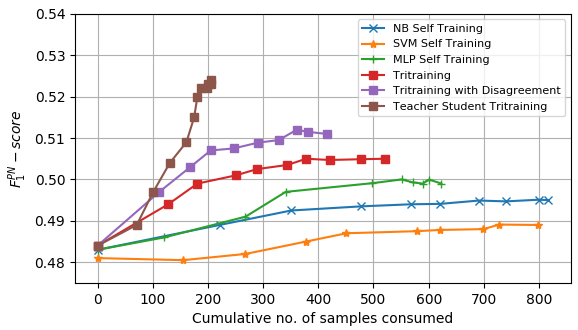}
    \caption{$F_1^{PN}$ score with cumulative number of samples used for all baselines for $40$\% label rate.}
    \label{fig:num_samples}
\end{figure}

\noindent{\bf Quantity of Unlabeled Data Consumed.} 
To evaluate the effectiveness of our stopping criterion, we calculate the quantity of unlabeled data consumed during the self-labeling process. Figure \ref{fig:num_samples} shows a plot of the models' $F_1^{PN}$ with regard to the cumulative number of samples added throughout the iterations (each datapoint in the plot corresponds to an iteration). We find that the proposed method consumes only 201 unlabeled instances to reach the best prediction performance, whereas both the original tri-training and tri-training with disagreement added around twice or thrice the number of samples. From Figure 1, we can further see that many of the baseline algorithms reach the saturation point way before they stop the training process i.e. the improvement in performance is marginal or even decays under some circumstances. This proves the effectiveness of the proposed stopping criteria.


We see that our approach performs worse than the tri-training baselines in the earlier iterations. This happens because our algorithm learns easier cases in the very beginning and gradually increases the difficulty along the learning process. On the contrary, the original tri-training grows very fast but also plateaus earlier, hence not achieving the full potential of using the three base classifiers. This early plateauing is avoided in our case with the adoption of the adaptive thresholds.

\noindent{\bf Sensitivity Analysis.} 
We further perform sensitivity analysis for the assessment of the initial settings of $\tau_t$ and $\tau_s$ with respect to their impact on the model performance. Specifically, we compare the experiment results with: (1) the initial teacher threshold $\tau_t$ set over $[0.7,1.0]$ with initial $\tau_s$ fixed as $0.85$; and (2) the initial student threshold $\tau_s$ set over $[0.6,0.95]$ with initial $\tau_t$ fixed as $0.94$. In both settings, $\tau_t$ and $\tau_s$ are continuously updated with the learned adaptive rates $\lambda_t$ and $\lambda_s$ after their initial assignment. We observe only marginal performance losses with an average difference of $-0.015$ $F_1^{PN}$ over all values.  This indicates that the initial value for $\tau_t$ and $\tau_s$ would not affect the performance that much, as long as they are adaptive.


\section{Conclusion}
\label{sec:conclusion}
In this paper, we propose a new teacher-student paradigm for original tri-training with continuously adaptive threshold and a natural stopping criteria. We show that our model outperforms all self-training and tri-training baselines in terms of achieving higher overall performance, higher quality of generated proxy labels, while consuming a less quantity of the unlabeled data. Although we only validate the proposed method against the benchmark SemEval dataset in this paper, our ultimate goal is to utilize it as a solution for the scenarios with limited labeled data and to tackle real-world problems, where labeled data is hard to find or expensive to attain.


\bibliographystyle{konvens2019}
\balance
\bibliography{konvens2019.bib}








\end{document}